%% file: main.tex
\documentclass[nohyperref]{article}

\usepackage{microtype}
\usepackage[pdftex]{graphicx}
\usepackage{subfigure}
\usepackage{booktabs} 
\usepackage{algorithm}
\usepackage{multirow}
\usepackage{makecell}

\usepackage[pagebackref,breaklinks,colorlinks]{hyperref}

\usepackage[accepted]{icml2022}

\usepackage{amsmath}
\usepackage{amssymb}
\usepackage{mathtools}
\usepackage{amsthm}

\usepackage[capitalize,noabbrev]{cleveref}

\theoremstyle{plain}

\theoremstyle{definition}

\theoremstyle{remark}

\usepackage[textsize=tiny]{todonotes}

\icmltitlerunning{Winning the Lottery Ahead of Time: Efficient Early Network Pruning}

\input{preamble}
\begin{document}

\twocolumn[
\icmltitle{Winning the Lottery Ahead of Time: Efficient Early Network Pruning}



\icmlsetsymbol{equal}{*}

\begin{icmlauthorlist}
\icmlauthor{John Rachwan}{tum}
\icmlauthor{Daniel Zügner}{tum}
\icmlauthor{Bertrand Charpentier}{tum}
\icmlauthor{Simon Geisler}{tum}
\icmlauthor{Morgane Ayle}{tum}
\icmlauthor{Stephan Günnemann}{tum}
\end{icmlauthorlist}

\icmlaffiliation{tum}{Technical University Munich, Germany}

\icmlcorrespondingauthor{John Rachwan}{john.rachwan@tum.de}

\icmlkeywords{Machine Learning, ICML}

\vskip 0.3in
]



\printAffiliationsAndNotice{} 

\newcommand\earlycrop{Early \textbf{C}ompression via G\textbf{r}adient Fl\textbf{o}w \textbf{P}reservation}
\newcommand\cropacr{CroP}
\newcommand\cropsacr{\cropacr-S}
\newcommand\cropuacr{\cropacr-U}
\newcommand\cropitacr{CroPit}
\newcommand\cropitsacr{\cropitacr-S}
\newcommand\cropituacr{\cropitacr-U}
\newcommand\earlycropacr{Early\cropacr}
\newcommand\earlycropsacr{\earlycropacr-S}
\newcommand\earlycropuacr{\earlycropacr-U}

\input{sec/0_abstract}
\input{sec/1_introduction}

\input{sec/3_related}

\input{sec/2_preliminaries.tex}
\input{sec/4_method}

\input{sec/5_results}

\input{sec/6_conclusions}





\bibliography{main}
\bibliographystyle{icml2022}

\input{sec/X_supplementary}


\end{document}

%% file: preamble.tex
\usepackage{overpic}
\usepackage{enumitem} 
\usepackage{overpic} 
\usepackage{color}

\definecolor{turquoise}{cmyk}{0.65,0,0.1,0.3}
\definecolor{purple}{rgb}{0.65,0,0.65}
\definecolor{dark_green}{rgb}{0, 0.5, 0}
\definecolor{orange}{rgb}{0.8, 0.6, 0.2}
\definecolor{red}{rgb}{0.8, 0.2, 0.2}
\definecolor{darkred}{rgb}{0.6, 0.1, 0.05}
\definecolor{blueish}{rgb}{0.0, 0.3, .6}
\definecolor{light_gray}{rgb}{0.7, 0.7, .7}
\definecolor{pink}{rgb}{1, 0, 1}
\definecolor{greyblue}{rgb}{0.25, 0.25, 1}

\usepackage{blindtext}

\renewcommand{\paragraph}[1]{\vspace{1em}\noindent\textbf{#1}.}

%% file: sec/0_abstract.tex
\begin{abstract}

\looseness=-1
Pruning, the task of sparsifying deep neural networks, received increasing attention recently. Although state-of-the-art pruning methods extract highly sparse models, they neglect two main challenges: (1) the process of finding these sparse models is often very expensive; (2) unstructured pruning does not provide benefits in terms of GPU memory, training time, or carbon emissions. We propose Early Compression via Gradient Flow Preservation (EarlyCroP), which efficiently extracts state-of-the-art sparse models before or early in training addressing challenge (1), and can be applied in a structured manner addressing challenge (2). This enables us to train sparse networks on commodity GPUs whose dense versions would be too large, thereby saving costs and reducing hardware requirements. We empirically show that EarlyCroP outperforms a rich set of baselines for many tasks (incl. classification, regression) and domains (incl. computer vision, natural language processing, and reinforcment learning). \earlycropacr{} leads to accuracy comparable to dense training while outperforming pruning baselines.

\end{abstract}

%% file: sec/1_introduction.tex
\section{Introduction}
\label{sec:intro}
State-of-the-art deep learning typically operates in the overparametrized regime. However, a large body of literature has shown that a high number of carefully chosen parameters can be removed (i.e. pruned) while maintaining the network's predictive performance \cite{lecun,molchanov,evci2020rigging, su2020sanitychecking,lee2018snip,Wang2020Picking}.
\input{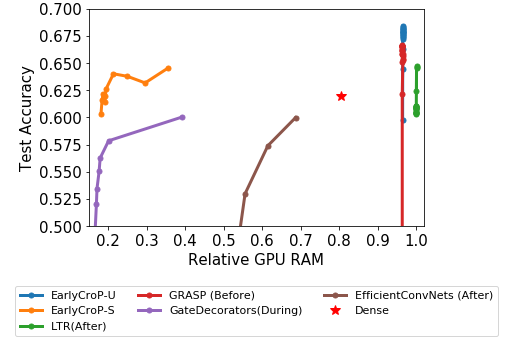}

\looseness=-1
It was first believed that sparse networks obtained from pruning pre-trained networks cannot be retrained from scratch. However, \cite{frankle2018the} presented the Lottery Ticket Hypothesis (LTH): randomly initialized deep neural networks contain sparse sub-networks (winning tickets) that -- when trained in isolation -- achieve test performance comparable to the fully trained dense model. This hypothesis suggests that we could prune a large number of a network's weights \emph{at initialization} (i.e., before training) and still obtain the full performance after training. That being said, the procedure proposed in \cite{frankle2018the} involves training the dense model to convergence multiple times, which is computationally very expensive. SNIP \cite{lee2018snip} and GraSP \cite{Wang2020Picking} were then proposed with the goal of pruning a randomly initialized model at initialization using a \textit{sensitivity criterion} for each weight.

\looseness=-1
Due to its decreased performance on large network/dataset combinations, the LTH was later revised for very deep networks. The authors note that for training the sparse model, we need to initialize its weights to the dense model's weights from a certain point \textcolor{black}{early} in training \cite{frankle2020linear}. This suggests that the best performing subnetworks can be found early in training (instead of before). Finding the earliest point at which we can prune without losing performance is challenging, and the authors present Linear Mode Connectivity (LMC), a computationally very expensive approach involving training multiple copies of the network. This suggests that to achieve a good trade-off between performance and efficiency of finding these subnetworks, pruning should follow the same spirit as before training methods but be applied \textit{early} instead.

\looseness=-1
Besides the question of \emph{when} to prune, an orthogonal dimension is structured vs.\ unstructured pruning. Unstructured pruning prunes individual weights (i.e., sets weight matrix elements to zero), while structured pruning removes entire neurons (i.e., rows/columns of weight matrices or convolution filters). Thus, structured pruning can reduce the training/inference time, memory footprint, and carbon emissions of the model; unstructured pruning has no significant impact on the above. On the other hand, structured pruning is generally much more challenging, and most previous works (including the LTH) perform unstructured pruning. 

\looseness=-1
Unless we maintain the learning dynamics of a neural network, pruning will hinder the learning process. The learning dynamics of a feed forward neural network can be described through the Neural Tangent Kernel (NTK) which approximately remains constant after some epochs of training in networks \cite{goldblum2020truth}. If we preserve the NTK while pruning, we expect the training process to not be affected. We develop a novel and principled pruning method that preserves the Gradient Flow (GF). By leveraging the close relation of the NTK to the GF, we show that we can prune a network while keeping the effect on the NTK minimal. Furthermore, we use the connection between GF and NTK to track when the learning dynamics become stable enough to perform \emph{early} pruning.


\looseness=-1
We present \earlycrop{} (\earlycropacr{}), a method for pruning a network early in training. \earlycropacr{} requires training the model only once, yet maintains the dense network's performance at high levels of sparsity. Thus, in the unstructured setting, \earlycropacr{} is about 5 times less expensive than the LTH. In addition, our method can be applied \emph{before} or \emph{early} in training, and extended to \emph{structured} pruning. Performing structured pruning before training provides a better accuracy/efficiency trade-off than most previous structured baselines, and enables us to train sparse networks \emph{whose dense versions would not fit into the GPU}. Furthermore, \earlycropacr{} reduces carbon emissions by up to 70\% without affecting dense performance and can thus help mitigate the environmental impact of deep learning and reduce training and inference costs.

\textbf{Contributions. }We approach neural network pruning with the explicit goal of unlocking real-world, practical improvements. Our key contributions are:

\begin{itemize}[leftmargin=*]
 \setlength\itemsep{-.1em}
    \item \textbf{Why to prune?} We transfer a GF based pruning criterion to be applicable for structured pruning, which allows faster forward and backward passes using less GPU memory and computational cost, \textcolor{black}{while surpassing baselines in test accuracy};
    \item \textbf{How to prune?} We leverage a connection between the NTK and GF by using a pruning criterion that aims to minimally affect the GF, and therefore the NTK and learning dynamics;
    \item \textbf{When to prune?} We further utilize the connection between GF and NTK to indicate the smooth transition to the \emph{lazy kernel regime}, the phase during which we can prune the network with little effect on the training dynamics. Thus, this brings the cost saving of structured pruning during training as well. \textcolor{black}{We also show that our method can be applied before training, reducing costs even further at only a small drop in accuracy.}

\end{itemize}

These contributions unlock substantial real-world benefits for practitioners and researchers: we can train large sparse models on commodity GPUs whose dense counterparts would be too large. We evaluate our approach extensively over a diverse set of model architectures, datasets, and tasks.

%% file: fig/teaser.tex
\begin{figure}[t]
\centering
\includegraphics[width=0.99\linewidth]{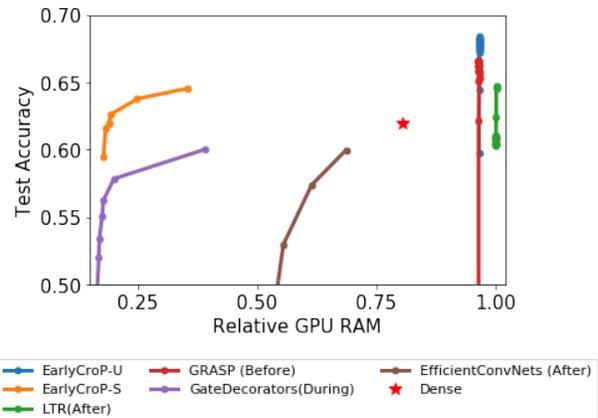}
\caption{
Our \earlycropsacr{} provides the best trade-off between GPU RAM consumption and test accuracy, outperforming the accuracy of the dense model and other structured pruning baselines. Unstructured baselines hardly bring benefits in GPU RAM or training time, but can achieve better test accuracy, with our \earlycropuacr{} reaching the highest values. Dataset: CIFAR-100, architecture: VGG16.
}
\label{fig:teaser}
\end{figure}

%% file: sec/3_related.tex
\section{Related work}
\label{sec:related}

\textbf{Pruning Criterion.} In order to prune network weights, they need to be ranked according to an importance score. This concept is not new, in fact, it was introduced in `Optimal Brain Damage' \cite{lecun} and `Optimal Brain Surgeon' \cite{hassibi}. Yet, it only regained traction when \cite{han} showed successful deep compression by pruning based on weight magnitude. Most pruning research since then has followed this approach \cite{NEURIPS2019_1113d7a7, evci2020rigging, mostafa2019parameter,bellec2018deep,dettmers2020sparse,mocanu2018scalable, You2020Drawing, chen2020lottery}. However, the biggest drawback of using weight magnitudes is that the network needs to be trained first to achieve a good accuracy. Therefore, more recent works have focused on scoring weights without the need for training using first order \cite{lee2018snip, tanaka2020pruning} and second order \cite{Wang2020Picking, lubana2021gradient-flow-framework} information. Note that the pruning process can be applied in one-shot or iteratively \cite{jorge2021progressive, verdenius2020pruning}.

\textbf{Pruning Time. }Up until the introduction of the LTH \cite{frankle2018the}, the consensus in the literature was that pruned models cannot be trained from scratch. Therefore, all sparse networks were extracted either from pre-trained networks \cite{han,lecun,hassibi, wangchaoqi, li2016pruning}, or throughout training \cite{srinivas2016generalized, louizos2018learning, evci2020rigging, mostafa2019parameter,bellec2018deep,dettmers2020sparse,mocanu2018scalable}. However, the LTH showed that there exist sparse models within the original randomly initialized dense model that can achieve comparable performance to the dense model. That being said, the LTH’s pruning algorithm, Iterative Magnitude Pruning (IMP), requires multiple iterations of a train-prune cycle. Nevertheless, the LTH's findings motivated works that strove to extract these sparse networks directly from the randomly initialized dense network \cite{su2020sanitychecking,lee2018snip,Wang2020Picking,dejorge2020progressive,frankle2020pruning,verdenius2020pruning}. 

The first method that introduced pruning before training was SNIP \cite{lee2018snip}, with the goal of preserving weights that have the highest effect on the loss. A subsequent work, GraSP \cite{Wang2020Picking}, uses the Hessian-gradient product in its score and prunes the weights with the goal of increasing the Gradient Flow (GF). Finally, \citet{lubana2021gradient-flow-framework} show that GraSP can lead to an increasing loss and instead propose to prune the weights that least affect the GF.

\looseness=-1
The performance of the LTH degrades with bigger networks and datasets \cite{frankle2020stabilizing}. Subsequently, the LTH was updated to indicate that the best performing sparse models do not necessarily exist at initialization but rather that they appear early in training. 

\looseness=-1
To the best of our knowledge, the only work that explores the extraction of sparse models early in training is Early Bird Tickets \cite{You2020Drawing}. They perform structured pruning early in training when the Hamming Distance between pruning masks at subsequent epochs becomes smaller than some threshold. However, they do not offer any theoretical justification for pruning early in training and they only show their results for a maximum pruning ratio of 70\%, suggesting that the Hamming Distance is not an effective criterion to achieve high sparsities.

\looseness=-1
\textbf{The Early Phase of DNN Training. }
Another line of work aims to analyze the early phase of neural network training. \citet{gur-ari2019gradient} study the Hessian eigenspectrum and observe that during training, a few large eigenvalues emerge in which gradient descent happens, whereas the rest get close to zero. However, these observations depend on the architecture. \citet{achille2019critical} found that the network goes through  critical training periods during which perturbing the data can cause irreversible damage to the network's final performance, after which the network becomes robust to these perturbations. However, the critical periods occur very late in the training process. Finally, \citet{frankle2020linear} propose the Linear Mode Connectivity (LMC) as a method for detecting when networks become stable to SGD noise. However, LMC is extremely expensive, requiring to train two copies of a network to completion at every epoch.

\looseness=-1
\textbf{Structured Pruning. }Pruning methods are divided into two categories: (1) unstructured methods which generate a binary mask that is applied before every forward pass \cite{frankle2018the,lee2018snip,tanaka2020pruning, Wang2020Picking}, and (2) structured methods that remove entire neurons or convolutional filters \cite{ding, li2016pruning, louizos2018learning, verdenius2020pruning, you2019gate}. Unstructured pruning is the more common variant for its simplicity and ease of implementation. However, since the entire dense network remains the same size, pruning does not provide improvements in \textcolor{black}{GPU RAM}, time, and carbon emissions. \textcolor{black}{While these improvements can be obtained for unstructured pruning by using operations on sparse compressed matrices, they require significant changes to the network when dealing with advanced layers.} Conversely, structured pruning \textcolor{black}{reduces the size of weight matrices, thereby requiring less space, time and energy during training and inference}. We highlight: (1) SNAP \cite{verdenius2020pruning}, which adapts the SNIP \cite{lee2018snip} score to the structured setting to prune before training, (2) Gate Decorators \cite{you2019gate}, which builds on top of \cite{liu2017learning} by adding a sensitivity-based criterion and pruning the network iteratively during training, and (3) EfficientConvNets \cite{li2016pruning}, which prunes a pre-trained network by scoring filters by their $L_1$-norm. \textcolor{black}{Other recent works include \cite{8816678, He_2019_CVPR, 10.1145/3295500.3356156} but we omit them since GateDecorators outperforms them.}

%% file: sec/2_preliminaries.tex
\section{Background}

\looseness=-1
\textbf{Neural Tangent Kernel (NTK). } 
The NTK is defined as $g_Y(\Theta_t)^Tg_Y(\Theta_t)$ \cite{jacot2018} where $g_Y(\Theta_t)$ denotes the gradient of the model prediction $Y$ w.r.t.\ the model parameters $\Theta_t$ at time $t$. The NTK is known to accurately describe the \emph{dynamics of the network's prediction during training}, under the assumption that the following Taylor expansion holds:
\begin{align}
\label{eq:taylor-prediction}
    Y(\Theta_t) \approx Y(\Theta_0) + g_Y(\Theta_0)^T(\Theta_t - \Theta_0)
\end{align}
Under the NTK assumption, a neural network reduces to a linear model with the Neural Tangent Kernel (NTK). The NTK assumption is particularly accurate for wide neural networks. In practice, this assumption holds (i.e.\ the NTK remains approximately constant) after the model training dynamic has transitioned from the \emph{rich active regime} to the \emph{lazy kernel regime} (see Section~\ref{sec:whentoprune}).

\looseness=-1
\textbf{Gradient Flow (GF).} We define the GF as $g_L(\Theta_t)^Tg_L(\Theta_t)$ \cite{lubana2021gradient-flow-framework}, where $g_L(\Theta_t)$ denotes the gradient of the model loss $L$ w.r.t.\ the model parameters $\Theta_t$. The GF is known to accurately describe the \emph{dynamics of the network's gradient norm during training}, under the assumption that the following Taylor expansion holds:
\begin{align}
    &g_L(\Theta_t)^Tg_L(\Theta_t) = ||g_L(\Theta_t)||^2_2 \\ 
    \label{eq:taylor-gradient-norm}
    &\approx ||g_L(\Theta_0)||^2_2 +2 (H_L(\Theta_0)g_L(\Theta_0))^T(\Theta_t - \Theta_0)
\end{align}
where $H_L(\Theta_t)$ denote the model's Hessian at time $t$. In order to prune the weights which least affect the GF, \citet{lubana2021gradient-flow-framework} propose to use the following weight importance score:
\begin{align}
\label{eq:weight-score}
    I(\Theta_t) = |\Theta_t^TH_L(\Theta_t)g_L(\Theta_t)|\,,
\end{align} 
and remove $\rho$\% of the parameters with the lowest scores. Preserving the GF stands in stark contrast to the importance score of GraSP~\cite{Wang2020Picking} $I_{\text{GraSP}}(\Theta_t) = -\Theta_t^TH_L(\Theta_t)g_L(\Theta_t)$, that maximizes the GF. Note that while the importance score \eqref{eq:weight-score} was initially used before training, we propose to use this importance score to prune during training either in one-shot or iteratively. 

%% file: sec/4_method.tex
\section{Method}

The core motivation of our work is to improve the applicability of sparse neural networks w.r.t.\ concrete real-world metrics such as carbon emissions, price, time or memory at both training and inference time. To this end, our method first transfers the pruning criterion \eqref{eq:weight-score} to structured pruning, thus allowing faster forward and backward passes (see Sec.~\ref{sec:whytoprune}). Second, we derive a relation between the NTK and the GF suggesting that preserving the GF also preserves the NTK (see Sec.~\ref{sec:howtoprune}). Hence, the pruning criterion \eqref{eq:weight-score} is a suitable importance weight score for pruning a neural network once the NTK assumption holds i.e.\ in the \emph{lazy kernel regime}. Third, our method detects when we enter the \emph{lazy kernel regime} to prune early in training without impacting the training dynamics \textcolor{black}{(see Sec.~\ref{sec:whentoprune})}, thereby extending the cost saving of our (structured) sparse neural networks to the training phase while achieving a high test accuracy.

\subsection{Why to prune?}
\label{sec:whytoprune}

\looseness=-1
The main use case of unstructured pruning is to highlight the overparametrized nature of neural networks. In particular, while dense-like sparsity for deep learning \cite{zhou_learning_2021} is a promising research direction, it suffers from multiple downsides: \textbf{(a)} models are typically only sparsified in the forward pass and hence dense-like sparsity has limited potential in speeding up training. \textbf{(b)} Not all deep learning frameworks (e.g.\ PyTorch) support it. \textbf{(c)} Only the newest GPUs (starting Nvidia Ampere 2020) support dense-like sparsity.

\looseness=-1
In order to really benefit from pruning, we need to \emph{prune full structures} (neurons and channels) instead. This reduction in dimensions/channels directly translates into lower computational cost on existing GPUs \emph{without} further implementation efforts or any usage of any specialized tensor operations. Thus, this leads to a sparse model that provides improvements in time, memory and carbon emissions. Combined with the fact that we can apply our pruning method before and early in training, we can drastically reduce not only model costs after training but during training as well (see Sec.~\ref{sec:whentoprune}). 

\looseness=-1
In order to achieve structured pruning, we need to score entire nodes instead of individual weights, i.e. generate a score for a node's activation function $f_l$. However, since $f_l$ is simply a function and not a learnable parameter, we cannot use pruning score \eqref{eq:weight-score} directly. Instead, similarly to \citet{verdenius2020pruning}, we define auxiliary gates of a layer $l$ by $c_l=1$ over each node's input, which in turn will act as a learnable parameter whose gradient information represents the activation's information. We can formally define this for a linear layer $l$ with weight $\Theta_l$ and bias $b_l$, and an input $x$ in the following way:
\begin{align}
    f_l(\Theta_l * x + b_l) &= f_l(c_l (\Theta_l * x + b_l))\\
    I(f_l) &= |H_L(c_l)g_L(c_l)|
\end{align}
After scoring the nodes using the auxiliary gates, the pruning process follows the original one by removing $\rho$\% of nodes.

\subsection{How to prune?}
\label{sec:howtoprune}

In this section, we draw an important connection between GF and NTK showing that pruning the weights with the lowest importance score \eqref{eq:weight-score} aims at preserving the training dynamics of both the network's gradient norm and the network's prediction during training. First, we observe that GF and NTK are connected by the following relation:
\begin{align}
\label{eq:gf-ntk-relation}
    GF &= g_L(\Theta_t)^Tg_L(\Theta_t) \\ 
    &= g_L(Y)^Tg_Y(\Theta_t)^Tg_Y(\Theta_t)g_L(Y)\\
    &= g_L(Y)^T NTK g_L(Y)
\end{align}
Second, \citet{lubana2021gradient-flow-framework} present evidence that preserving the GF also implicitly preserves the model loss $L(\Theta_t)$. In particular, preserving the GF also preserves the gradient of the loss w.r.t.\ the prediction $g_L(Y)$. Hence, the relation \eqref{eq:gf-ntk-relation} and the preservation $g_L(Y)$ imply that the NTK is also preserved when the GF is preserved. 

Furthermore, given that Taylor expansions \eqref{eq:taylor-prediction} and \eqref{eq:taylor-gradient-norm} hold, the pruning criteria \eqref{eq:weight-score} which preserves the GF -- which maintains the gradient-norm dynamics -- is also preserving the NTK -- which maintains the prediction dynamics. This remark is crucial since while the dynamics of the neural network's predictions during training can be approximated well by \eqref{eq:taylor-prediction} during the \emph{lazy kernel regime}, this approximation might not be accurate during the \emph{rich active regime}.

\subsection{When to prune?}\label{sec:whentoprune}

First, we show in an introductory experiment that the pruning time has an important impact on the final accuracy of the pruned model. Hence, we train multiple ResNet50 models on CIFAR100 and prune each to $98\%$ from epoch $0$ (i.e. before pruning) to epoch $80$ (see Fig.\ref{fig:prunepoint}). We observe that \textbf{(1)} the longer we train the dense model before pruning, the higher the final accuracy of the sparse model, and most importantly \textbf{(2)} after a certain point in time, further training of the dense model before pruning does not bring significant improvement on the final accuracy. Indeed, we observe  a 3\% improvement in the final accuracy of the sparse model when pruning at epoch 1 of training instead of before training, an 11\% improvement when pruning at epoch 26, and no great improvement when pruning after epoch 30.

\looseness=-1
We now introduce the pruning time detection score used by \earlycropacr{} which is motivated from both practical and theoretical perspectives. \earlycropacr{} aims to detect the best time for pruning in two steps: \textbf{(1)} at every epoch we compute the pruning time score 
\begin{align} \label{pruningscore}
    \Delta^t_0 = ||\Theta(t)-\Theta(0)||^2,
\end{align}
and, \textbf{(2)} if the difference of the scores at two subsequent epochs relative to the initial score $\Delta^{1}_0$ is smaller than a defined threshold $th$,
\begin{align}\label{scorediff1}
\frac{|\Delta^t_0 - \Delta^{t-1}_0|}{|\Delta^{1}_0|} < th
\end{align} 
we run the \earlycropacr{} pruning algorithm described in Algorithm \ref{alg:crop}. It can be clearly deduced that the smaller the pruning time score, the more negligible the second order term in eq.  \ref{eq:taylor-gradient-norm} will be, making the latter a good approximation. Additionally, by the triangle inequality, we can extract the following upper bound from \ref{scorediff1}
\begin{align}\label{scorediff}
\frac{|\Delta^t_0 - \Delta^{t-1}_0|}{|\Delta^{1}_0|} \leq \frac{\|\Theta(t)-\Theta(t-1)\|}{\|\Theta(1)-\Theta(0)\|}
\end{align}
which is expected to lie in $[0, 1]$ when weights change less significantly over time. Hence, the scale of the threshold is expected to be similar for different models and datasets. The complexity of computing the score is $\mathcal{O}(m)$ where $m$ is the number of model parameters, thus incurring only minor computation overhead at every epoch to detect the pruning time.

\looseness=-1
From a theoretical perspective, the pruning time detection algorithm's goal is to detect when the linearization of the prediction dynamics \eqref{eq:taylor-prediction} assumed by the NTK holds during training. In the early phase of training called \emph{rich active regime}, neural network parameters move to a significant distance from the initial weights. Thus the linearization of the prediction dynamics \eqref{eq:taylor-prediction} usually does not hold in early training epochs and the NTK quickly changes. This \emph{rich active regime} is crucial to achieve high performance, in particular for deep models \cite{woodworth2020regime}. In the second phase of training called \emph{lazy kernel regime}, the parameters move by a small distance, thus making the linearization \eqref{eq:taylor-prediction} a good approximation of the training dynamics of the predictions  \cite{sun2019optimization}. Since our importance weight score \eqref{eq:weight-score} assumes the linearization \eqref{eq:taylor-prediction}, the best moment to prune is when the model transitions to the \emph{lazy kernel regime} during which the NTK is approximately constant. Further, previous works \cite{sun2019optimization, amari2020target, ghorbani2020linearized} showed that constancy of the NTK is a consequence of a constant weight norm during training. The transition to the lazy kernel training regime \textcolor{black}{is gradual} and can be detected when the relative change in the weight norm from initialization $\Delta^t_0$ becomes roughly constant i.e. when $\frac{|\Delta^t_0 - \Delta^{t-1}_0|}{|\Delta^{1}_0|}$ becomes very close to 0. 

\looseness=-1
In practice, we expect the pruning time criteria to be a reliable indicator of the final accuracy of the pruned model. Indeed, we observed that the detection score correlates well with the final test accuracy of the sparse model (see Fig.\ref{fig:prunepoint}). It can be clearly seen that the smaller the detection score at the moment of pruning, the higher the final test accuracy of the pruned model. In practice, we observed in Fig.~\ref{fig:prunepoint} and in Fig.~\ref{fig:additionalprunepoint} in the appendix that pruning to higher sparsities benefits more from longer training. Therefore, we use $th=1-\rho$ which connects the time pruning threshold $th$ to the target sparsity $\rho$. As desired, a higher target sparsity $\rho$ leads to longer dense training (see Fig.\ref{fig:additionalprunepoint}).

%% file: sec/5_results.tex
\input{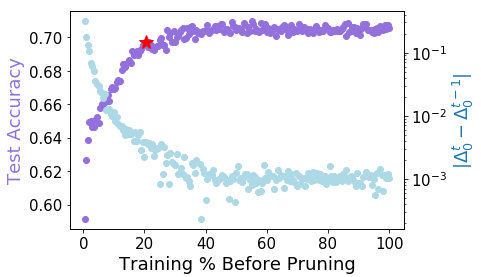} 
\section{Empirical Evaluation}

We now show the effectiveness of our \earlycropacr{} for structured pruning (\earlycropsacr{}) and unstructured pruning (\earlycropuacr{}). For this we determine the point for early pruning as described in Section \ref{sec:whentoprune}. We use \cropacr{} as our pruning criterion before training and \cropitacr{} if pruning is additionally performed iteratively. \textcolor{black}{We use a cloud instance of GTX 1080 TIs for all experiments}. Further details about the experimental setup can be found in the appendix. The code and further supplementary material is available online\footnote{\url{www.cs.cit.tum.de/daml/early-crop/}}.
\input{fig/imageclassification}

\looseness=-1
\textbf{Image Classification.} The datasets used for Image Classification are the common public benchmarks CIFAR10 \cite{Krizhevsky09}, CIFAR100 \cite{Krizhevsky09}, and Tiny-Imagenet \cite{5206848}. Regarding networks, we use ResNet18, VGG16, ResNeXt-101 32x16d and ResNeXt-101 32x48d. For unstructured pruning baselines, we use random pruning, SNIP \cite{lee2018snip}, GraSP \cite{Wang2020Picking}, and LTR \cite{frankle2020linear}. For structured pruning baselines, we use random pruning, EfficientConvNets \cite{li2016pruning}, GateDecorators \cite{you2019gate}, and SNAP \cite{verdenius2020pruning}. \textcolor{black}{All models are trained for 80 epochs, except LTR which re-trains the network up to 10 times}. We report train and test accuracy, weight and node sparsity, 
batch and total training time in seconds, GPU memory in GB, disk size in MB, carbon emitted \textcolor{black}{from the extraction and training of the sparse model} using CodeCarbon \cite{codecarbon} in grams. Note that total training time includes the time to find and train the sparse model.

\looseness=-1
\textbf{Regression.} We evaluate a Fully Convolutional Residual Network \cite{laina2016deeper} on the NYU Depth Estimation task \cite{Silberman:ECCV12}. We compare \earlycropsacr{} and \earlycropuacr{} against all unstructured baselines since they are stronger than the structured baselines. All pruned models are trained for 10 epochs. We report the performance using the Root Mean Squared Error (RMSE).

\looseness=-1
\textbf{Natural Language Processing (NLP).} 
We evaluate the Pointer Sentinal Mixture Model\cite{Merity2017PointerSM} on the PTB language modeling dataset\cite{PTBDATASET}. We compare \earlycropsacr{} and \earlycropuacr{} to all unstructured baselines since they are stronger than the structured baselines. We train the pruned models for 30 epochs and we report the achieved log perplexity.

\textbf{Reinforcement Learning (RL).} 
We use the FLARE framework \cite{akbik2019flair} to evaluate a simple 3-layer FCNN with layer size 256 using the A2C algorithm on the classic control game CartPole-v0 \cite{openaigym}. We run 20 agents with 640 games each. We compare our \earlycropsacr{} and \earlycropuacr{} against LTR and Random baselines. All pruned models are trained for 30 epochs. We report the performance of the pruned models using the average returned environment reward.

\subsection{Image Classification}

\textbf{Accuracy.} We present the accuracy over different sparsity levels for the model-dataset combinations ResNet18/ CIFAR10, ResNet18/Tiny-Imagenet, VGG16/CIFAR10, and VGG16/CIFAR100 in Figure~\ref{fig:imageclassification} a-d, respectively. Our methods \earlycropsacr{} and \earlycropuacr{} consistently outperform all other methods but the LTR, where we perform on par. However, as will be discussed later, the LTR comes with a 3-5 times higher training time than the dense model while our methods reduce training time. There are two further exceptions when it comes to the best accuracy on ResNet18/Tiny-ImageNet. First, GateDecorators performs as well as \earlycropsacr{}. Second, for lower sparsity rates, \earlycropuacr{} is outperformed by some methods that prune before training.

\textbf{Structured vs.\ Unstructured.} \earlycropsacr{} closes the accuracy gap of existing approaches between unstructured and structured pruning on CIFAR10 dataset. However, a gap remains for the larger and more complex datasets CIFAR100 and Tiny-Imagenet. Nevertheless, structured pruning can be used to reduce the training time and memory requirements. This also implies that with our \earlycropsacr{} we can use a larger model while saving compute (see Sec.~\ref{sec:largemodel}).

\input{tab/resnet18cifar10new}
\input{tab/vgg16cifar10}
\textbf{Training cost.} In Tables ~\ref{table:resnet18cifar10} \& ~\ref{table:vgg16cifar10} we complement the accuracy with the training time, batch time, GPU RAM, Disk space and CO$_2$ emissions for a sparsity of 95\% on ResNet18/CIFAR10 and 98\% on VGG16/CIFAR10 respectively. We see that our \earlycropsacr{} is not only preserving the high accuracy but also comes with significant improvement in training time (33\% and 36\% resp.) and time per batch (39\% and 61\% resp.). It is as efficient as the other structured pruning methods or outperforms them. When only considering the CO$_2$ footprint, \cropitsacr{} and SNAP outperform methods that prune later in training. In the appendix, we also give details about different model-dataset combinations. In summary, the stated observations also hold for the other evaluated model-dataset combinations.

\textbf{Pruning early vs.\ before.} Pruning early in training (i.e. when we enter the \emph{lazy kernel regime}) outperforms pruning before training. Our \earlycropsacr{} and \earlycropuacr{} have a clear edge over the methods that prune before training. This is even true if we use the same pruning criterion (\cropacr{}) and additionally prune iteratively (\cropitacr{}). The only drawback of pruning early in training vs.\ before is that for the first epochs we either require more GPU RAM or need to reduce the batch size. For the model size on disk we do not find a significant difference among pruning methods.

\subsection{Pruning a Large Model}\label{sec:largemodel}
The goal of this experiment is two-fold: we show that \textbf{(1)} our criterion can be used to prune large models that don't fit on commodity GPUs, 
and \textbf{(2)} the resulting sparse model matches the performance of the dense one, and outperforms a dense model of the same size. To this end, we introduce the ResNext101\_32x48d as our large model, a network that has 829 Million parameters and requires 15.5 GB to be loaded into GPU memory; exceeding the memory of common GPUs such as the RTX 3080 Ti. Nevertheless, with our method we can still efficiently train such a large model. For this, we perform one initial pruning step before training using the CPU and then continue on a commodity GPU as usual. We also introduce the ResNext101\_32x16d as our smaller dense model which has 193 Million parameters and requires 3.9 GB to be loaded into GPU memory. The results of the experiment are depicted in Table \ref{table:resnext}. 


First, we observe that the large ResNext101\_32x48d pruned to 98.5\% weight sparsity matches the performance of its dense counterpart in test accuracy. Moreover, training the sparse subnetwork has a 14 times smaller carbon footprint, is 7 times faster to train, is 192 times smaller on disk, and takes 4.9 times less GPU memory than the large dense model. Interestingly, the pruned model also outperforms the ResNext101\_32x16d model of the same size, while training 6.2 times faster and emitting 9.5 times less carbon. Finally, we show that when training for more epochs, the sparse model achieves an even bigger performance gap compared to both dense models while still taking less total training time. This experiment not only shows that \cropsacr{} makes training large models on commodity machines possible, but can extract sparse models that are more efficient and more accurate compared to dense models of the same size.


\input{tab/resnext}
\input{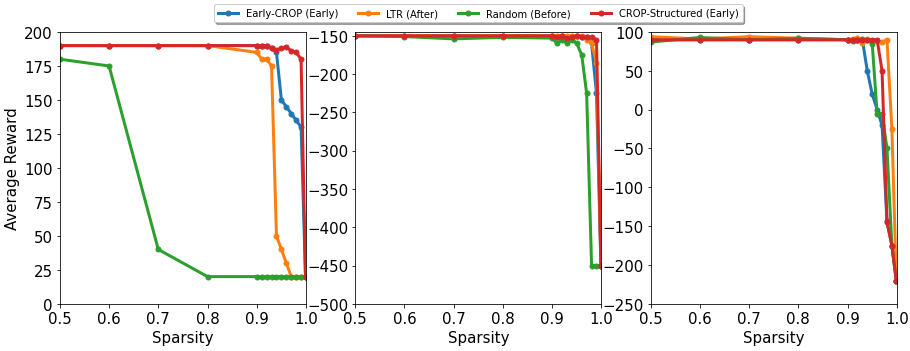}
\subsection{Regression}
Regarding Regression, we can see from Figure \ref{fig:Regression} that both variants of \earlycropacr{} preserve the dense model's RMSE even at 99.9\% weight sparsity. All Before training methods except GraSP have an instant increase to 0.20 RMSE with continuous decline at higher sparsities. Surprisingly, Random Pruning outperforms GraSP at all pruning ratios. This is due to GraSP pruning entire layers and limiting the network's learning capabilities.
\input{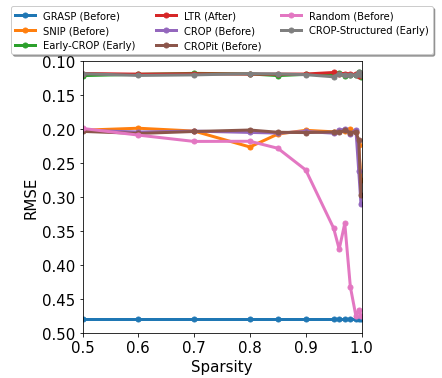}

\subsection{Natural Language Processing}
NLP presents itself as the most challenging out of all evaluated tasks. Nevertheless, both versions of \earlycropacr{} outperform all other baselines until 89\% sparsity (see Figure \ref{fig:NLP}). Beyond that, the unstructured version is on par or slightly better than other unstructured baselines whereas the structured version continues to outperform all compared baselines. In this task, the importance of early pruning is accentuated by the large gap between the early and before versions of \cropacr{}. Interestingly, LTR performs very poorly compared to all other baselines on all reported pruning ratios. Indeed, certain layers in the PSMM network converge to small weight magnitudes during training compared to the rest of the network. This means that any pruning method relying solely on the weight magnitudes, and operating at a global scale in the network, would prune these layers entirely, leading to an untrainable network. Thus, this experiment highlights the importance of gradient-based information when evaluating the importance of model parameters. We show additional NLP results by evaluating BERT on multiple language tasks (see Appendix \ref{sec:nlp}).
\input{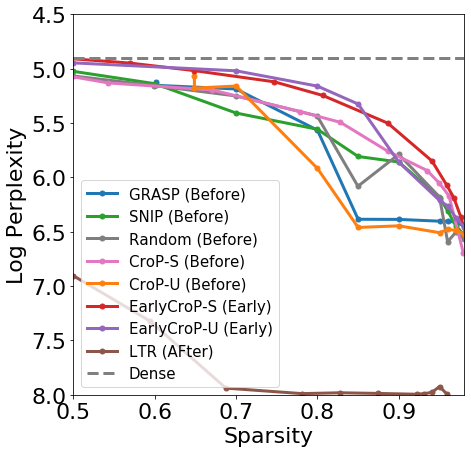}
\subsection{Reinforcement Learning}
We can observe from Figure \ref{fig:RL} that \earlycropacr{} outperforms LTR in both the structured and unstructured setting.
Note that the \earlycropsacr{} once again outperforms its unstructured counterpart. However, if we allow the unstructured models to train for a longer time, they achieve similar performance to the structured version. This can be explained by the ease of training of structured models, which are still fully-connected models where all computed gradients contribute to the weight updates, whereas unstructured models compute gradients that are not used by the pruned weights, rendering the training slower.

%% file: fig/prunepoint.tex
\begin{figure}[h]
\includegraphics[width=1\linewidth]{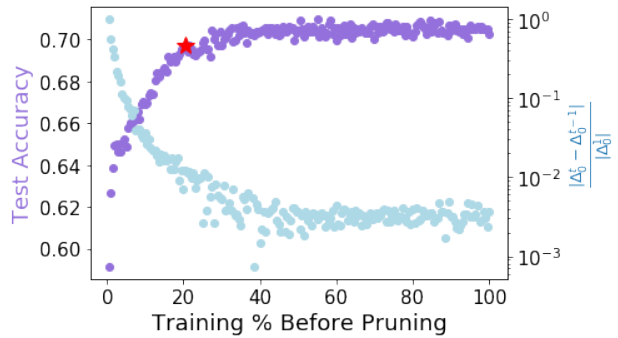}
\centering
\vskip 0.1in
\caption{ResNet50 pruned to sparsity 98\% at different points during training on CIFAR100 plotted against the final test accuracy of the pruned model after 200 epochs of training (purple) and the difference between the Relative Weight Change in two subsequent epochs (blue). \textcolor{red}{$\bigstar$} denotes when the Relative Weight Change in two subsequent epochs is below the threshold $th = 1-\rho$.}
\label{fig:prunepoint}
\centering
\end{figure}

%% file: fig/imageclassification.tex
\begin{figure*}
\begin{center}
\includegraphics[width=\textwidth]{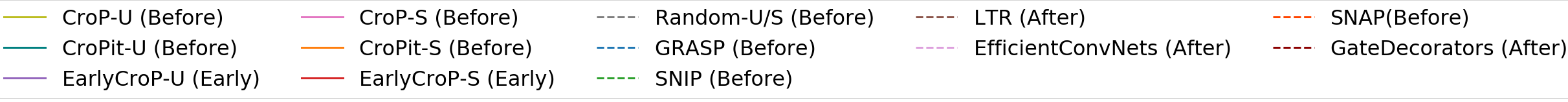}
\subfigure[]{\includegraphics[width=\columnwidth]{fig/resnet18cifar10-figure.pdf}}
\hfill
\subfigure[]{\includegraphics[width=\columnwidth]{fig/resnet18tiny-figure.pdf}}
\subfigure[]{\includegraphics[width=\columnwidth]{fig/vgg16cifar10-figure.pdf}}
\hfill
\subfigure[]{\includegraphics[width=\columnwidth]{fig/vgg16cifar100-figure.pdf}}

\end{center}
\caption{Structured (Left) and Unstructured (Right) test accuracy for ResNet18/CIFAR10 (a), ResNet18/Tiny-Imagenet (b), VGG16/CIFAR10 (c), and VGG16/CIFAR100 (d) with increasing weight sparsity.
}
\label{fig:imageclassification}
\end{figure*}

%% file: tab/resnet18cifar10new.tex
\newcommand{\STAB}[1]{\begin{tabular}{@{}c@{}}#1\end{tabular}}
\begin{table*}[!htbp]
\caption{Comparison between different pruning criteria on ResNet18/CIFAR10 at 95\% sparsity, averaged over three runs. Since for structured pruning we cannot precisely control the weight sparsity, we show results closest to 95\% weight sparsity. $\pm$ denotes standard deviation, and $\uparrow$/$\downarrow$ indicate metrics where higher/lower is better. Bold/underline indicate best/second best results. GPU RAM and Disk correspond to those of the final pruned model.}
\vskip 0.15in
\centering
\resizebox{0.9\textwidth}{!}{\begin{tabular}{clrrrrrrrr}
\toprule
  & \textbf{Method} &  \textbf{Test accuracy} $\uparrow$ &  \makecell{\textbf{Weight}\\ \textbf{sparsity}} &  \makecell{\textbf{Node}\\ \textbf{sparsity}} & \makecell{\textbf{Training}\\\textbf{time (h)}} $\downarrow$ & \makecell{\textbf{Batch}\\\textbf{time (ms)}}$\downarrow$ & \makecell{\textbf{GPU RAM}\\\textbf{(GB)}} $\downarrow$   &  \makecell{\textbf{Disk}\\\textbf{(MB)}} $\downarrow$ & \makecell{\textbf{Emissions}\\\textbf{(g)} }$\downarrow$ \\
\midrule
 & Dense & 91.5\% {\scriptsize
 $\pm$ 0.12} & - &   - &   0.78 & 109 & 2.38 &   398 & 83\\
\midrule
\multirow{8}{*}{\STAB{\rotatebox[origin=c]{90}{Structured}}} & 
Random-S & 86.3\% {\scriptsize $\pm$ 0.06} & 93.7\% &   75.0\% &   0.68 &  82 &   0.62 &   24.9 & 38 \\
& SNAP & 87.6\% {\scriptsize $\pm$ 0.94} & 93.6\% &   72.6\% &   0.70 &   81 &   0.63 &   25.4 & 39 \\
& \cropsacr{} & 87.5\% {\scriptsize $\pm$ 0.36} & 93.6\% &   72.3\% &  0.67 &  91 & 0.63 &  25.4 & 43 \\
& \cropitsacr{} & \underline{87.8\%} {\scriptsize $\pm$ 0.33} & 95.0\% &  74.5\% &  \underline{0.52}& \textbf{64} &  0.59 & 19.6 & 35 \\
\cmidrule{2-10}
& EarlyBird & 84.3\% {\scriptsize $\pm$ 0.32} & 95.3\% & 65.0\% & \textbf{0.48} & 72 & 0.58 & 19.1& 55\\
& \earlycropsacr{} & \textbf{91.0\%} {\scriptsize $\pm$ 0.52} & 95.1\% & 65.8\% &  \underline{0.52} & \underline{66} & 0.56 & 19.2 & 68 \\ 
\cmidrule{2-10}
& GateDecorators & 87.3\% {\scriptsize $\pm$ 0.09} & 95.7\% &   73.7\% &  0.72 &  83 & 0.58 &   17.2 & 54 \\
\cmidrule{2-10}
& EfficientConvNets & 70.5\% {\scriptsize $\pm$ 0.53} & 95.9\% & 79.7\% & 0.77  & 83 &  0.76 & 25.4 & 63 \\
\midrule
\multirow{8}{*}{\STAB{\rotatebox[origin=c]{90}{Unstructured}}} & Random-U & 84.9\% {\scriptsize $\pm$ 0.24} & 95.0\% &  - &   0.78 &  \underline{102} &   2.86 &   12.0 & 79 \\
& SNIP & 88.2\% {\scriptsize
 $\pm$ 0.57} & 95.0\% &   - &   0.79 & 105 & 2.86 &   12.0 & 80 \\
& GRASP & 88.4\% {\scriptsize $\pm$ 0.13} & 95.0\% &   - &   0.79 & 106 & 2.84 & 12.0 & 81 \\
& \cropuacr{} & 87.9\% {\scriptsize $\pm$ 0.16} & 95.0\% &   - &   \underline{0.75} &  107 &   2.88 &   12.0 & 79 \\
& \cropituacr{} & 89.1\% {\scriptsize $\pm$ 0.24} & 95.0\% &   - &   0.80 &  113 &   2.88 &   12.0 & 87 \\ 
\cmidrule{2-10}
& \earlycropuacr{} & \underline{91.1}\% {\scriptsize $\pm$ 0.23} & 95.0\% &  - &  \textbf{0.74} &  \textbf{97} &  2.86 & 12.0 & 83\\
\cmidrule{2-10}
& LTR & \textbf{91.5\%} {\scriptsize $\pm$ 0.26} & 95.0\% & - &  1.94 &   111 &  2.51 & 12.0 & 202 \\
\bottomrule
\end{tabular}}
\label{table:resnet18cifar10}
\end{table*}

%% file: tab/vgg16cifar10.tex
\begin{table*}[!htbp]
\vskip 0.2in
\caption{Comparison between different pruning criteria on VGG16/CIFAR10 at 98\% sparsity. Since for structured pruning we cannot precisely control the weight sparsity, we show results closest to 98\% weight sparsity. $\uparrow$/$\downarrow$ indicate metrics where higher/ lower is better. Bold/ underline indicate best/ second best results. \textcolor{black}{GPU RAM and Disk correspond to those of the final pruned model.}}
\vskip 0.2in
\centering
\resizebox{0.9 \textwidth}{!}{\begin{tabular}{clrrrrrrrr}
\toprule
  & \textbf{Method} &  \textbf{Test accuracy} $\uparrow$ &  \makecell{\textbf{Weight}\\ \textbf{sparsity}} &  \makecell{\textbf{Node}\\ \textbf{sparsity}} & \makecell{\textbf{Training}\\\textbf{time (h)}} $\downarrow$ & \makecell{\textbf{Batch}\\\textbf{time (ms)}}$\downarrow$ & \makecell{\textbf{GPU RAM}\\\textbf{(GB)}} $\downarrow$   &  \makecell{\textbf{Disk}\\\textbf{(MB)}} $\downarrow$ & \makecell{\textbf{Emissions}\\\textbf{(g)} }$\downarrow$ \\
\midrule
- & Dense & 90.2\%& - & - &   1.82 &  290 & 1.02  & 1720  & 246\\
\midrule
\multirow{8}{*}{\STAB{\rotatebox[origin=c]{90}{Structured}}} & 
Random-S & 89.3\%& 98.0\% & 86.1\% &  \textbf{0.67} &  \textbf{82} &  0.23 &  33.6 & 43 \\
& SNAP & 89.8\% & 98.2\% &   89.0\% &   \underline{0.68} &  \underline{89}  &  0.22 &  30 & 55\\
& \cropsacr{} & 91.1\% & 98.0\% & 88.0\% &     0.71 &  91    & 0.23 &  33.6 & 83 \\
& \cropitsacr{} & \underline{92.4}\% & 98.0\% &  88.0\% &  0.81 &  112  &  0.23 & 30.4 & 100 \\
\cmidrule{2-10}
& EarlyBird & 85.9\%  & 98\% & 89 \% & 0.52 & 110 & 0.32 & 36.2& 160\\

& \earlycropsacr{} & \textbf{93.0}\% & 98.0\% & 89.0\% & 1.16 &  112 & 0.63 & 36.0 & 156 \\ 
\cmidrule{2-10}
& GateDecorators & 90.0\%  & 98.0\% & 87.0\%  & 1.07 &  111  & 0.23  & 37.8  & 143 \\
\cmidrule{2-10}
& EfficientConvNets & 84.2\%  & 98.0\% & 86.0\% & 1.66  & \underline{89} &  0.64 & 34.2 & 209 \\
\midrule
\multirow{8}{*}{\STAB{\rotatebox[origin=c]{90}{Unstructured}}} & Random-U & 88.5\% & 98.0\% &  - &  2.03  &  159 &   1.22 &  35.0 &  247\\
& SNIP & 90.1\% & 98.0\% &   - &  \underline{2.02}  & 157 & 1.22 &  35.0 & 248 \\
& GRASP & 92.0\% & 98.0\% &  - &  2.03 & 157 & 1.23 & 35.0 & 249 \\
& \cropuacr{} & 91.8\% & 98.0\% &   - &  \underline{2.02} &  157&   1.22 &   35.0 &  248\\
& \cropituacr{} & 91.6\%& 98.0\% &   - & \underline{2.02} &  157&  1.22 &  35.0 & 249\\ 
\cmidrule{2-10}
& \earlycropuacr{} & \underline{93.0}\%& 98.0\% &  - & \textbf{2.01}&  157 &  1.22 & 35.0 & 250 \\
\cmidrule{2-10}
& LTR & \textbf{93.6}\% & 98.0\% & - & 4.07  & 158   &  1.22 & 35.0 & 592 \\
\bottomrule
\end{tabular}}

\label{table:vgg16cifar10}
\end{table*}

%% file: tab/resnext.tex
\begin{table}[h!]
\caption{Comparison of a pruned ResNext101\_32x48d (RN48) model and a similar sized dense ResNext101\_32x16d (RN16) model (CIFAR10). RN48-S are models pruned with \cropsacr{}.}
\vskip 0.2in
\centering
\resizebox{\columnwidth}{!}{
\begin{tabular}{rrrrrrrrr}
\toprule
\textbf{Model} &  \makecell{\textbf{Test}\\\textbf{acc.}} &  \makecell{\textbf{Weight}\\\textbf{sparsity}} &  \makecell{\textbf{Node}\\\textbf{sparsity}} & \textbf{Epochs} & \makecell{\textbf{Training}\\\textbf{time (h)}} &  \makecell{\textbf{VRAM}\\\textbf{(GB)}} & 
\makecell{\textbf{Emissi-}\\\textbf{ons (g)}} \\
\midrule
RN48 &  92.4\% &  - &- & 30 &  4.60 & 18.84  &  634	 \\
RN16 &  92.1\% &  - &- & 30 &  4.02 & 3.89  &   445\\
RN48-S &  \textbf{92.5\%} &  98.5\% & 89.9\% & 30 &  \textbf{0.64}  &  3.56  &   47 \\
\midrule
RN48-S &  93.2\% &  98.5\% &89.9\% & 80 &  2.60 & 3.56  &   194 \\
\bottomrule
\end{tabular}}

\label{table:resnext}
\end{table}

%% file: fig/rl.tex
\begin{figure}[h]
\vskip 0.3in
\includegraphics[width=0.99\linewidth]{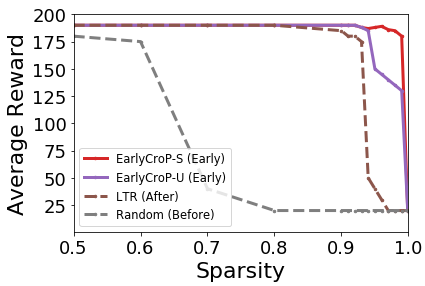}
\centering
\vskip 0.1in
\caption{Sparse model performance on the classic control game Cartpole-v0 in average reward (higher is better)}
\label{fig:RL}
\centering
\end{figure}

%% file: fig/regression.tex
\begin{figure}[h]
\includegraphics[width=0.99\linewidth]{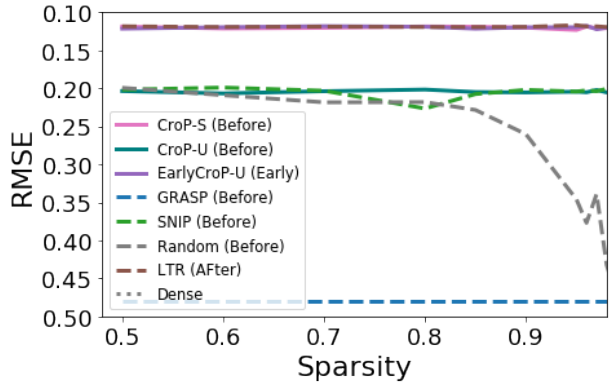}
\centering
\vskip 0.1in
\caption{Sparse model performance on the NYU Depth Estimation task in RMSE (lower is better)}
\label{fig:Regression}
\centering
\end{figure}

%% file: fig/nlp.tex
\begin{figure}[h]
\vskip 0.3in
\includegraphics[width=0.99\linewidth]{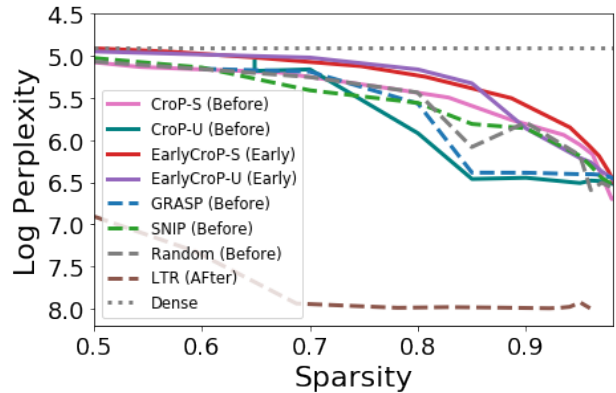}
\centering
\vskip 0.1in
\caption{Sparse model performance on the PTB language modelling task in log perplexity (lower is better)}
\label{fig:NLP}
\centering
\end{figure}

%% file: sec/6_conclusions.tex
\section{Conclusion}

\looseness=-1
We have demonstrated that, for vision, NLP, and RL tasks, \earlycropuacr{} extracts winning tickets matching and often outperforming those found by the LTR by pruning early in training when the model enters \emph{lazy kernel training}. Additionally, we showed that \earlycropsacr{} outperforms other structured methods, providing the best trade-off between final test accuracy and efficiency in terms of time, space, and carbon emissions. Finally, we show that we can use \cropsacr{} to train models that do not fit on commodity GPUs by extracting sparse models that preserve the initial model's performance and outperform a similarly sized dense model for the same number of epochs. Thus, our methods bring tangible real-world benefits for researchers and practitioners. We hope that the results shown in this paper motivate more research on the study of structured pruning in the early phase of DNN training.

%% file: sec/X_supplementary.tex
\appendix

\setcounter{page}{1}

\twocolumn[
\centering
\Large
\vspace{0.5em}Supplementary Material \\
\vspace{1.0em}
] 
\appendix

\section{Training and Inference Cost Computation}
This section details the computations in Figure \ref{fig:teaser}. We introduce the V100 16GB GPU (2.48\$/h) and the V100 32GB GPU (4.96\$/h). We use the total training time needed to train RN48 and RN48-S for 30 epochs each (see Table \ref{table:resnext}. Consequently, in order to train RN48, we need $4.96\$/h \times 4.6 h = 22.816\$$. In order to train RN48-S, we need $2.48\$/h \times 0.64h = 1.59\$$
\section{Algorithm}
\input{Algorithm/earlycrop}

\section{Experimental Setup}
\subsection{Optimization}
\textbf{Image Classification. }For all experiments, we use the ADAM\cite{adamoptimizer} optimizer and a learning rate of $2e{-3}$. The One Cycle Learning Rate scheduler is used to train all models except VGG16. The batch size used for CIFAR10 and CIFAR100 experiments is 256 while for Tiny-Imagenet it is 128. All sparse models are allowed to train the same amount of epochs (80) to converge which, except for LTR, includes the number of epochs required to extract the sparse model. In the case of LTR the final sparse model is allowed to train for 80 epochs.\\
\textbf{Regression. }For all experiments, we use a batch size of 8, the ADAM optimizer with a learning rate of $1e-5$. All pruned models are trained for 10 epochs.
\\
\textbf{Natural Language Processing. }For all experiments, we use a batch size of 128, and the ADAM optimizer with a learning rate of $1e-3$. All pruned models are trained for 30 epochs.
\\
\textbf{Reinforcement Learning. }A description of the models used and number of runs used for each environment can be found in Table \ref{table:rl}.
\input{tab/rl}
\subsection{Datasets Pre-Processing}
\paragraph{CIFAR10 \cite{Krizhevsky09}}
We augment the normalized CIFAR10 with Random Crop and Random Horizontal Flip. Images are additionally resized to $64\times64$.

\paragraph{CIFAR100 \cite{Krizhevsky09}}
We augment the normalized CIFAR100 with Random Crop, Random Horizontal Flip, and Random Rotation.

\paragraph{Tiny-Imagenet \cite{5206848}}
We normalize the dataset and resize each image to $224 \times 224$.

\section{Evaluation Metrics}
In this section we describe how specific metrics are calculated.

\paragraph{Time}
We report time in two separate ways. First, we report the total time required (Training time), which is defined from the start of the experiment until the sparse model's training is finished. Second, we report the time it takes to perform a full forward and backward pass (Batch time) on a given batch using the CUDA time measurement tool \cite{NEURIPS2019_9015}.

\paragraph{GPU RAM}
The RAM footprint of a process refers to how much memory it consumes on the GPU. This effectively includes the costs of loading the model and performing a training step on it. We use the CUDA memory measurement tool to report this metric\cite{NEURIPS2019_9015}.

\paragraph{Disk Storage}
We estimate the storage needed to store a model on disk using the CSR sparse matrix format \cite{10.1145/1583991.1584053}. Similarly to \cite{verdenius2020pruning}, we used a ratio of 16:1 float precision on all vectors of the CSR format.

\paragraph{Energy Emissions}\label{emissions}
We estimate CO$_2$ emissions in g using CodeCarbon emissions tracker \cite{codecarbon}. These estimates consider all emissions from the start of experiments until the end of training.

\section{Additional Results}
\subsection{Reinforcement Learning}
See Figure \ref{fig:lunar} for experiments on the Acrobot-v1 and LunarLander-v2 environments.
\input{fig/rl_appendix}
\subsection{Natural Language Processing}\label{sec:nlp}
We evaluated BERT on multiple language tasks (see Table \ref{table:bert}). At the same pruning sparsity, EarlyCroP-U outperforms LTR on 5 out of 8 tasks while training 10× faster.
\subsection{Pruning Point Experiments}
In Figure \ref{fig:additionalprunepoint} we present more experiments on pruning models at different points in training. We can clearly observe a correlation between the desired pruning rate and the optimal time to prune. The higher the desired final sparsity, the longer the network should be trained before being pruned.
\input{fig/additionalprunepoint}

\subsection{VGG16/CIFAR100}
See Table \ref{table:vgg16cifar100} for comparisons between different pruning criterions at the same pruning level on VGG16/CIFAR100.
\input{tab/vgg16cifar100}

\subsection{ResNet18/TinyImageNet}
See Table \ref{table:resnet18tiny} for comparisons between different pruning criterions at the same pruning level on ResNet18/Tiny-Imagenet.
\input{tab/resnet18tinyimagenet}
\subsection{VGG16/ImageNet}
In Table \ref{tab:imagenet} we present a comparison between a dense VGG16 and a sparse VGG16 pruned using EarlyCroP-S on the ImageNet-2012 (ILSVRC2012) \cite{5206848} classification dataset. Given 62 hours of training on ImageNet-2012 (ILSVRC2012) on a signle V100 GPU, EarlyCroP-S on VGG16 (50\% weights pruned) achieves an accuracy of 61.43\% while the dense model only achieves 58.78\%. Moreover, for the same number of training epochs, EarlyCrop-S achieves 60.01\% in 51 hours while the dense model achieves 58.78\% in 62 hours.
\input{tab/imagenet}

\input{tab/bert}

%% file: Algorithm/earlycrop.tex
\begin{algorithm}
\caption{Early-CroP}\label{alg:crop}
\begin{algorithmic}[1]
 \STATE Initialize the weights $\Theta$, weight mask $Mask_0$ with all 1s, pruning ratio $\rho$, maximum dense training time $m$, score threshold $th$, and the number of pruning iterations $it$;
 \STATE $\Theta_0 = \Theta$
 \STATE $\Theta_{prev} = \Theta$
\WHILE{$t(epoch) < m$}
 \STATE Do one SGD epoch;
 \STATE diff = $\frac{||\Theta(t)-\Theta(0)||^2}{||\Theta(0)||^2}$ - $\frac{||\Theta(t-1)-\Theta(0)||^2}{||\Theta(0)||^2}$ 
\IF{diff $\leq th$}
      \STATE break;
\ENDIF
    \STATE $\Theta_{prev} = \Theta$
\ENDWHILE
\WHILE{$i \leq it$}
      \STATE $\rho_i = \rho_{final} - (\rho_{final}-1/2)\times(1/2)^i$ 
      \STATE $I(\Theta) = |\Theta_t^TH_L(\Theta_t)g_L(\Theta_t)|$
      \STATE $Mask_i=CroP(I(\Theta), p_i)$ 
     \STATE  Apply Mask on Network
\ENDWHILE
  \STATE Apply Mask on Network
\WHILE{$t\leq t_{max}$}
  \STATE Do one SGD epoch;
\ENDWHILE
\end{algorithmic}
\end{algorithm}

%% file: tab/rl.tex
\begin{table}[h!]
\caption{A summary of the setup used in Reinforcement Learning}
\vskip 0.15in
\centering
\resizebox{\columnwidth}{!}{\begin{tabular}{|l|r|r|r|r|}
\toprule
Name &  Network &  Algorithm &  Agents & Games \\
\midrule
                    CartPole-v0 &            MLP(128-128-128-out) &       A2C  & 16&8000\\                       \hline
                    Acrobot-v1 &            MLP(256-256-256-out) &           A2C   & 16   &8000 \\
                       \hline
                    LunarLander-v2 &            MLP(256-256-256-out) &          A2C    &16&8000\\

\bottomrule
\end{tabular}}
\label{table:rl}
\end{table}

%% file: fig/rl_appendix.tex
\begin{figure*}
\begin{center}
\includegraphics[width=\textwidth]{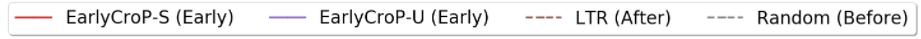}
\subfigure[]{\includegraphics[width=0.8\columnwidth]{fig/rl2.pdf}}
\hspace{1cm}
\subfigure[]{\includegraphics[width=0.8\columnwidth]{fig/lunalander.pdf}}
\hfill
\end{center}
\caption{Sparse model performance on classic control games (a) Acrobot-v1 and (b) LunarLander-2
}
\label{fig:lunar}
\end{figure*}

%% file: fig/additionalprunepoint.tex
\begin{figure*}
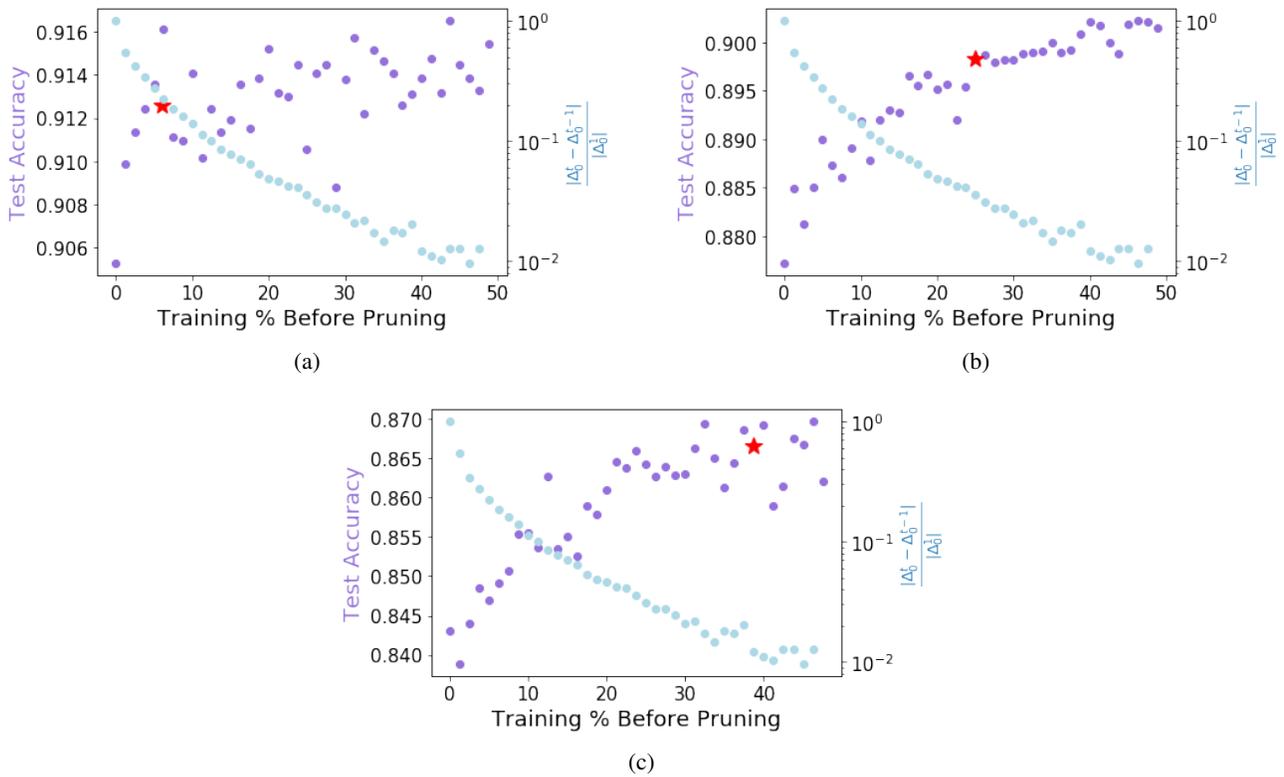

\begin{center}
\subfigure[]{\includegraphics[width=\columnwidth]{fig/prunepoint80-figure.pdf}}
\hfill
\subfigure[]{\includegraphics[width=\columnwidth]{fig/prunepoint98-figure.pdf}}
\hfill
\subfigure[]{\includegraphics[width=\columnwidth]{fig/prunepoint995-figure.pdf}}
\end{center}
\caption{ResNet18 pruned to sparsity 80\%(a), 98\%(b), and 99.5\% (c) at different points during training on CIFAR10 plotted against the final test accuracy of the pruned model after 80 epochs of training (purple) and the difference between the Relative Weight Change in two subsequent epochs (blue). All pruned models are trained for 80 epochs. \textcolor{red}{$\bigstar$} denotes when the Relative Weight Change in two subsequent epochs is below $th < 1-\rho$
}
\label{fig:additionalprunepoint}
\end{figure*}

%% file: tab/vgg16cifar100.tex
\begin{table*}[!htbp]
\caption{Comparison between different pruning criteria on VGG16/CIFAR100 at 98\% sparsity. Since for structured pruning we cannot precisely control the weight sparsity, we show results closest to 98\% weight sparsity. $\uparrow$/$\downarrow$ indicate metrics where higher/ lower is better. Bold/ underline indicate best/ second best results. \textcolor{black}{GPU RAM and Disk correspond to those of the final pruned model.}}
\vskip 0.15in
\centering
\resizebox{0.9 \textwidth}{!}{\begin{tabular}{clrrrrrrrr}
\toprule
  & \textbf{Method} &  \textbf{Test accuracy} $\uparrow$ &  \makecell{\textbf{Weight}\\ \textbf{sparsity}} &  \makecell{\textbf{Node}\\ \textbf{sparsity}} & \makecell{\textbf{Training}\\\textbf{time (h)}} $\downarrow$ & \makecell{\textbf{Batch}\\\textbf{time (ms)}}$\downarrow$ & \makecell{\textbf{GPU RAM}\\\textbf{(GB)}} $\downarrow$   &  \makecell{\textbf{Disk}\\\textbf{(MB)}} $\downarrow$ & \makecell{\textbf{Emissions}\\\textbf{(g)} }$\downarrow$ \\
\midrule
- & Dense &  62.1\%  & - &   - & 0.77 &   114 & 1.03 & 1745 &  88 \\
\midrule
\multirow{8}{*}{\STAB{\rotatebox[origin=c]{90}{Structured}}} & 
Random-S & 53.9\%& 98.0\% & 86.0\% &  \textbf{0.59} & 53 &  0.23 &  35 & 29 \\
& SNAP & 49.3\%  & 98.0\% &   89.0\% &   0.67  &  54 &  0.16 &  36 & 33 \\
& \cropsacr{} & \underline{57.4}\%  & 98.0\% & 89.0\% &  \underline{0.61} & \underline{46}  & 0.23 &  36 & 35 \\
& \cropitsacr{} & 56.5\%  & 98.1\% &  89.0\% & 0.62 & \textbf{44} &  0.23 & 33& 30 \\
\cmidrule{2-10}
& EarlyBird & 60.7\% & 98.0\% & 89.0\% & 0.56 & 68 & 0.20 & 36& 62\\
& \earlycropsacr{} & \textbf{62.2}\%  & 97.9\% & 88.0\% & 0.64  & 69 & 0.23 & 36& 58 \\ 
\cmidrule{2-10}
& GateDecorators & 55.0\%  & 97.9\% & 87.0\%  & \underline{0.61} &  78 & 0.23  & 36 & 68 \\
\cmidrule{2-10}
& EfficientConvNets & 29.5\%  & 98.0\% & 86.0\% &  0.72 & 55 &  0.24 & 36& 83 \\
\midrule
\multirow{8}{*}{\STAB{\rotatebox[origin=c]{90}{Unstructured}}} & Random-U & 55.8\% & 98.0\% &  - &  0.74  &  118 &  1.23 &  35 & 99 \\
& SNIP & 61.9\%  & 98.0\% &   - &  0.79  & 109 & 1.24 &  35 & 90 \\
& GRASP & 63.4\%  & 98.0\% &  - &  0.79 & 113 & 1.24 & 35 & 91 \\
& \cropuacr{} & 63.8\%  & 98.0\% &   - &  \textbf{0.74}  & 109  &   1.23 &  35 & 94 \\
& \cropituacr{} & 56.3\%  & 98.0\% &   - &   \textbf{0.74} & 111 &  1.23 &  35 & 91 \\ 
\cmidrule{2-10}
& \earlycropuacr{} & \textbf{65.1}\%  & 98.0\% &  - &  \textbf{0.74} &  109 &  1.23 & 35 & 91 \\
\cmidrule{2-10}
& LTR & \underline{64.7}\%  & 98.0\% & - &  3.44 &  109 &  1.28 & 35 & 301 \\
\bottomrule
\end{tabular}}

\label{table:vgg16cifar100}
\end{table*}

%% file: tab/resnet18tinyimagenet.tex
\begin{table*}[!htbp]
\caption{Comparison between different pruning criteria on ResNet18/TinyImageNet at 90\% sparsity. Since for structured pruning we cannot precisely control the weight sparsity, we show results closest to 90\% weight sparsity. $\uparrow$/$\downarrow$ indicate metrics where higher/ lower is better. Bold/ underline indicate best/ second best results. \textcolor{black}{GPU RAM and Disk correspond to those of the final pruned model.}}
\vskip 0.15in
\centering
\resizebox{0.9 \textwidth}{!}{\begin{tabular}{clrrrrrrrr}
\toprule
  & \textbf{Method} &  \textbf{Test accuracy} $\uparrow$ &  \makecell{\textbf{Weight}\\ \textbf{sparsity}} &  \makecell{\textbf{Node}\\ \textbf{sparsity}} & \makecell{\textbf{Training}\\\textbf{time (h)}} $\downarrow$ & \makecell{\textbf{Batch}\\\textbf{time (ms)}}$\downarrow$ & \makecell{\textbf{GPU RAM}\\\textbf{(GB)}} $\downarrow$   &  \makecell{\textbf{Disk}\\\textbf{(MB)}} $\downarrow$ & \makecell{\textbf{Emissions}\\\textbf{(g)} }$\downarrow$ \\
\midrule
- & Dense & 51.3\% &  - & - & 7.26  & 320 &  3.53 & 569 &   882\\
\midrule
\multirow{8}{*}{\STAB{\rotatebox[origin=c]{90}{Structured}}} & 
Random-S & 37.3\%  & 91.2\% & 80.0\% & 6.23 & 289 &  1.08 &  51 & 464 \\
& SNAP & 38.3\%  & 90.4\% &   82.6\% & \textbf{6.06}  &  268 &  0.84 &  55 & 514 \\
& \cropsacr{} & \underline{39.1}\%  & 90.1\% & 77.7\% & 6.72  & 237  & 1.11 &  54 & 615 \\
& \cropitsacr{} & \underline{39.1}\%  & 91.4\% &  79.3\% & 6.66 & 236 &  1.08 & 49 & 591 \\
\cmidrule{2-10}
& \earlycropsacr{} & \textbf{39.2}\%  & 90.8\% & 84.1\% & 7.03 & \underline{202} & 0.25 & 49 & 676 \\ 
\cmidrule{2-10}
& GateDecorators & 30.1\%  & 89.2\% & 91.2\%  & \underline{6.20} & \textbf{193} & 0.87  & 61  & 930 \\
\cmidrule{2-10}
& EfficientConvNets & 27.7\%  & 91.0\% & 79.8\% & 6.60  & 226 &  0.22 & 52 & 769 \\
\midrule
\multirow{8}{*}{\STAB{\rotatebox[origin=c]{90}{Unstructured}}} & Random-U & \underline{49.3}\%  & 90.0\% &  - &   7.25 &  351 &  4.20 &  57 & 932 \\
& SNIP & 46.2\%  & 90.0\% &   - &  7.27  & 314 & 4.18 &  57 & 854 \\
& GRASP & 43.7\%  & 90.0\% &  - & 7.27 & 315 & 4.22 & 57& 881 \\
& \cropuacr{} & 46.7\%  & 90.0\% &   - & 7.26  & 314  & 4.22 &   57& 877 \\
& \cropituacr{} & 19.1\%  & 90.0\% &   - &  7.26 & 313 &  4.22 &  57 & 890 \\ 
\cmidrule{2-10}

& \earlycropuacr{} & \textbf{49.8}\%  & 90.0\% &  - & 7.26 &  314 &  4.23 & 57 & 880 \\
\cmidrule{2-10}
& LTR & 46.3\%  & 90.0\% & - &  44.7 &  603 &  3.68 & 57 & 5540 \\
\bottomrule
\end{tabular}}

\label{table:resnet18tiny}
\end{table*}

%% file: tab/imagenet.tex
\begin{table*}[!htbp]
\small
\centering
\caption{EarlyCroP-S results on VGG16/ImageNet at 50\% weight sparsity}
\resizebox{0.9 \textwidth}{!}{\begin{tabular}{cccccccc}
\toprule
\multirow{4}{*}{} & Method&Top-1 Accuracy& Top-5 Accuracy&Train Time (hours)&Epochs& Batch Time (seconds) & GPU Memory (GB)\\
\cline{2-8}
&Dense& 58.78\% & 82.55\% & 62 & 18 & 1.01 & 12.15\\
&EarlyCroP-S& \textbf{61.43}\% & \textbf{87.01}\%& 62 & 26 & \textbf{0.66} & \textbf{10.63}\\
&EarlyCroP-S& 60.01\% & 83.38\%& 51 & 18 & \textbf{0.66} & \textbf{10.63}\\
\bottomrule
\end{tabular}}
\label{tab:imagenet}
\end{table*}

%% file: tab/bert.tex
\begin{table*}[!htbp]
\small
\centering
\caption{Comparison between LTH and EarlyCroP-U on different learnable BERT tasks.}
\resizebox{0.9 \textwidth}{!}{\begin{tabular}{ ccccccccccc } 
 \toprule
  \multirow{6}{*} & & MNLI & QQP & STS-B & WNLI & QNLI & RTE & SST-2 & CoLA & Training time \\
 \cmidrule{2-11}
  & Dense BERT & 82.39 & 90.19 & 88.44 & 54.93 & 89.14 & 63.30 & 92.12 & 54.51 & 1x\\
\cmidrule{2-11}
 & Sparsity & 70\% & 90\% & 50\% & 90\% & 70\% & 60\%& 60\% & 50\% & \\
 \cmidrule{2-11} 
 & LTR (Rewind 0)\% & 82.45 & 89.20 & 88.12 & 54.93 & 88.05 & \textbf{63.06}& 91.74 & 52.05 & 10x\\
 & LTR (Rewind 50)\% & \textbf{82.94} & 89.54 & \textbf{88.41} & 53.32 & 88.72 & 62.45 & 92.66 & 52.00 & 10x\\
 \cmidrule{2-11}
 & EarlyCroP-U & 82.11 & \textbf{89.99} & 88.02 & \textbf{56.33} & \textbf{89.12} & 62.1 & \textbf{92.03} & \textbf{52.2} & \textbf{1x}\\
 \bottomrule
\end{tabular}}
\label{table:bert}
\end{table*}